\pdfoutput=1
\documentclass[11pt]{article}

\usepackage[preprint]{acl}

\usepackage{times}
\usepackage{latexsym}

\usepackage[T1]{fontenc}
\usepackage[utf8]{inputenc}

\usepackage{microtype}

\usepackage{inconsolata}

\usepackage{graphicx}

\usepackage{graphicx}
\usepackage{tabularx}
\usepackage{booktabs}
\usepackage{adjustbox}
\usepackage{fontawesome}
\usepackage{amstext}
\usepackage{amsmath}
\usepackage{amssymb}
\usepackage{microtype}
\usepackage{multirow}
\usepackage{multicol}
\usepackage{algpseudocode}
\usepackage{amsfonts}
\usepackage{makecell}
\usepackage{colortbl} 
\usepackage{marvosym}
\usepackage{pifont}

\title{DeepSolution: Boosting Complex Engineering Solution Design via Tree-based Exploration and Bi-point Thinking}

\author{
Zhuoqun Li${}^{1,2}$,
Haiyang Yu${}^{3}$,
Xuanang Chen${}^{1}$,
Hongyu Lin${}^{1}$,
Yaojie Lu${}^{1}$,
  \\
{\bf ~Fei Huang${}^{3}$},
{\bf Xianpei Han${}^{1}$},
{\bf Yongbin Li${}^{3}$},
{\bf Le Sun${}^{1}$}
  \\
${}^{1}$Chinese Information Processing Laboratory,
Institute of Software, Chinese Academy of Sciences\\
${}^{2}$University of Chinese Academy of Sciences \\
${}^{3}$Tongyi Lab \\
{\tt \{lizhuoqun2021,chenxuanang,hongyu,luyaojie\}@iscas.ac.cn} \\
{\tt \{xianpei,sunle\}@iscas.ac.cn} \\
{\tt \{yifei.yhy,f.huang,shuide.lyb\}@alibaba-inc.com} \\
}

\begin{document}
\maketitle

\begin{abstract}
% This paper constructs SolutionBench to evaluate systems on complex engineering solution design, and propose SolutionRAG to generate reliable solutions via a bi-point thinking tree.

% Designing solutions for complex engineering requirements plays a critical role in human production activities. 
% However, previous research in the RAG field has not thoroughly concerned on  complex engineering solution design tasks. 
% To bridge this gap, we introduce a new benchmark, SolutionBench, to assess whether a system can generate complete and feasible solutions for engineering requirements with multiple complex constraints. 
% To further enhance complex engineering solution design, we propose SolutionRAG, a new system that utilizes tree-based exploration and bi-point thinking to generate reliable solutions. 
% The experimental results demonstrate that SolutionRAG is an effective system for complex engineering solution design, which is with the SOTA performance on the SolutionBench.

Designing solutions for complex engineering challenges is crucial in human production activities. 
However, previous research in the retrieval-augmented generation (RAG) field has not sufficiently addressed tasks related to the design of complex engineering solutions. 
To fill this gap, we introduce a new benchmark, SolutionBench, to evaluate a system’s ability to generate complete and feasible solutions for engineering problems with multiple complex constraints. 
To further advance the design of complex engineering solutions, we propose a novel system, SolutionRAG, that leverages the tree-based exploration and bi-point thinking mechanism to generate reliable solutions. 
Extensive experimental results demonstrate that SolutionRAG achieves state-of-the-art (SOTA) performance on the SolutionBench, highlighting its potential to enhance the automation and reliability of complex engineering solution design in real-world applications.
\url{https://github.com/Li-Z-Q/DeepSolution}.

\end{abstract}
\section{Introduction}

% 1. 针对复杂的工程需求进行方案设计是生产活动中的一项重要工作。
Designing solutions for complex engineering requirements is a crucial work in human production activities~\cite{ogot2004engineering,elmaraghy2012complexity}.
% 2. 需求中通常存在各种现实的限制条件，并且期待一个完整可行的解决方。例如，Please design a hospital construction solution to ensure the safety of personnel and equipment，forarea with annual rainfall of 3000 millimeters, expansive soil conditions, and badly frequent seismic activity. 
These requirements typically include multiple real-world constraints and expect complete and feasible solutions (e.g., \textit{Design a safe and efficient hospital construction plan in an area with annual rainfall of 3000 millimeters, expansive soil conditions, and frequent seismic activity}).
% 3. 人类专家完成此类工作的过程需要在查阅大量专业知识的基础上仔细斟酌方案从而确保满足复杂的需求，这一过程需要耗费大量的时间和人力。
Human experts complete such work by consulting extensive professional knowledge, carefully designing, and strictly deliberating, which require significant time and human resources~\cite{kalogerakis2010developing,de2011engineering}. 
% 4. 随着LLM和RAG技术的不断发展和进步，各工程领域都期待一种可靠的RAG系统来自动完成方案设计类工作。
With the continuous development of retrieval-augmented generation (RAG) techniques, the engineering fields anticipate a credible RAG system that can automatically generate reliable solutions for these complex engineering requirements~\cite{yu2024auto,zhou2024boosting}.

\begin{figure}[t!]
\centering
\includegraphics[width=\linewidth]{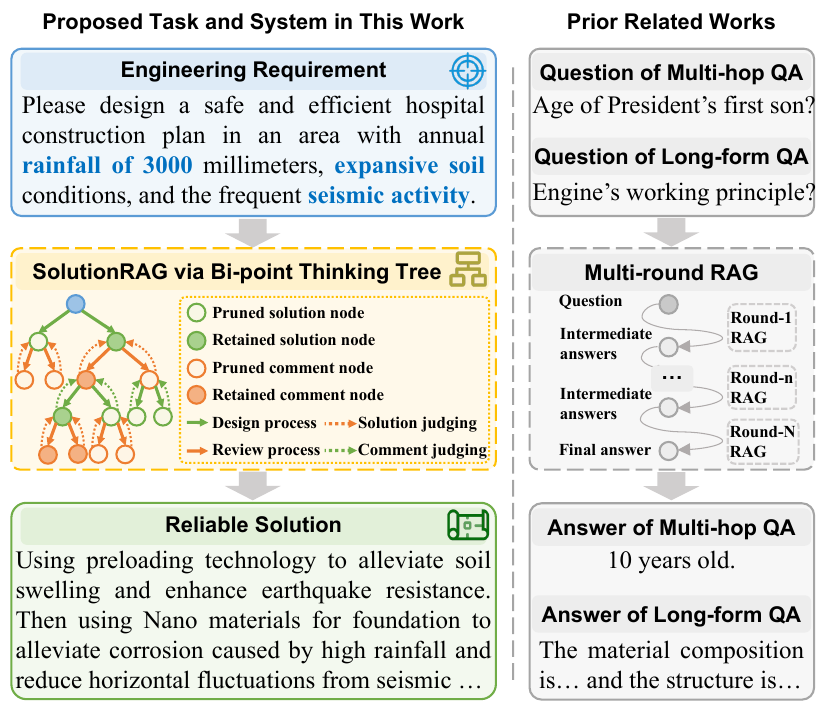} 
\caption{This paper proposes the complex engineering solution design task and a new system that can generate reliable solutions via the bi-point thinking tree.} 
\label{fig:head}
\end{figure}

% 1. Unfortunately，此前RAG领域中并没有对针对复杂的工程需求的方案设计类工作进行深入研究。
Unfortunately, prior works in RAG field do not sufficiently research the complex engineering solution design task.
% 2. 此前任务主要关注Multi-hop QA或者Long-form QA ，其问题是多个子问题的组合或者综合(e.g. 总统的夫人的家乡人口数量? 发动机的原理是什么? )，期待的回答是一个实体片段或者知识段落的拼接和整合(e.g., "3000人", "发动机的原理包括：能源输入...，动力输出...，齿轮结构...")。
Existing relevant papers mainly focus on Long-form QA or Multi-hop QA~\cite{zhu2024fanoutqa,tan2024proxyqa}, where the questions are integrated or composed of multiple sub-questions and the expected answers are typically assembled knowledge paragraphs or entity fragments.
% 3. 与这些任务不同，针对复杂的工程需求的方案设计类工作的需求中通常存在各种现实的限制条件，并且期待一个完整可行的解决方案。
Unlike these tasks, requirements of the complex engineering solution design task involve multiple real-world constraints and demand complete and feasible solutions~\cite{fortus2005design, jonassen2006everyday}, as shown in Figure~\ref{fig:head}. 
% 因此，基于RAG技术对于针对复杂的工程需求的方案设计类工作进行研究是一个有重要价值的亟需填补的空白。
Therefore, researching complex engineering solution design based on RAG  technology is a valuable gap that needs to be filled.

% 1. To fill this gap，我们首先提出了一个新的数据集CESDBench，用来考察RAG系统是否能够针对需求中通常存在各种现实的限制条件的需求得到一个完整可行的解决方案。
To fill this gap, we first introduce a new benchmark, \textbf{SolutionBench}, to evaluate whether a system can generate complete and feasible solutions for  complex engineering requirements with multiple constraints.
% 2. 首先，我们从各工程领域的权威期刊中收集针对复杂的工程需求的方案设计的技术报告
Firstly, to ensure the data source's authority, authenticity, and diversity, we collect thousands of  engineering reports about solution design from authoritative journals in various engineering domains. 
% 3. 然后，基于手工设计的模版利用LLM对技术报告进行内容抽取，获得源于现实的复杂需求、源于专家的权威解决方案、形成该方案时用到的对复杂需求进行解读的分析性知识和处理复杂需求时用到的技术性知识、以及专家设计该方案时的具体思路解读
Then, to build data that is convenient for testing and evaluation, we refer to the generative information extraction technologies~\cite{lu2022unifiedstructuregenerationuniversal,zhang-etal-2025-survey} and employ LLMs to extract useful content from these reports based on a manually formatted template, capturing real-world complex requirements, expert-authored solutions, analytical knowledge used to interpret the requirements, technical knowledge applied in addressing the requirements, and explanations for the expert's solution design process.
% 4. 最后，经过人工对LLM抽取结果的仔细校对之后，将同一领域下的所有知识合并为一个知识库，从而得到面向complex engineering solution design 的涵盖多种工程领域的benchmark
Finally, we manually verify and revise the extracted content, merge all knowledge within the same domain into a unified knowledge base, and then harvest a complete benchmark for complex engineering solution design that covers eight engineering domains.

% 1. 为了进一步深入研究上述针对复杂的工程需求的方案设计任务，我们提出SolutionRAG, which 通过双向思考和广泛探索生成可靠的方案
To further advance complex engineering solution design, we propose \textbf{SolutionRAG}, which can generate reliable solutions through tree-based exploration and bi-point thinking.
% 2. 首先，生成解决方案的思维路径并非只有一种，单链条思考可能会错过最佳推理路径。因此，方案生成时基于树结构进行广泛探索，每一个分支代表一个思考路径。
Firstly, the improvement process from suboptimal solutions to reliable solutions is flexible, rather than with a fixed reasoning pattern. Therefore, SolutionRAG conducts the tree-based exploration, where each branch represents a different improvement direction.
% 3. 第二，由于需求中存在各种现实的限制条件，单向思维的RAG不能保证生成的solution满足全部条件。因此，我们在树生长过程中交替进行方案Design和Review，从而逐渐提升方案的可靠性。
Secondly, due to the presence of multiple real-world constraints within the requirements, system-generated solutions cannot guarantee the satisfaction of all constraints. Therefore, SolutionRAG employs the bi-point thinking, which alternates between solution designing and reviewing during the tree growth, gradually improving  reliability of generated solutions.
% 4. 最后，为了提升推理效率，我们根据节点得分进行剪枝。每一个子节点都会对父节点进行打分，即review node判断上一层的design node的可靠性有多大，design node判断上一层的review node的帮助性有多大，经过这样交替打分，可以使推理过程保持在最有希望的方案和提升作用最大的审核意见构成的路径上，从而兼顾推理效率和性能。
Finally, to balance inference efficiency and performance, SolutionRAG implements pruning based on node evaluation, which can keep the inference process along the most promising solutions and the most helpful reviewed comments.

% In experiments, 我们基于SolutionBench考察了多种方法针对复杂的工程需求进行方案设计的能力，包括不进行RAG的深度思考模型、普通的RAG方法、多轮迭代RAG方法，以及我们提出的SolutionRAG。
In experiments, we evaluate various types of methods on SolutionBench to assess their ability in complex engineering solution design, including deep thinking models without RAG, standard RAG approaches, multi-round iterative RAG methods, and our SolutionRAG.
% 实验结果表明LLM仅依靠内部知识无法有效解决此类任务，此前的RAG方法也不能有效解决，而我们提出的SolutionRAG相对来讲是一种更有效的方案。
Experimental results show that LLMs relying solely on internal knowledge cannot effectively solve such tasks. Previous RAG methods also fail to generate satisfactory solutions. In contrast, our proposed SolutionRAG proves to be a more advanced approach. The main contributions of this paper can be summarized as follows:

\begin{itemize}
\setlength{\itemsep}{2pt}
\setlength{\parsep}{2pt}
\setlength{\parskip}{2pt}

\item We construct SolutionBench, which can evaluate a system's ability for complex engineering solution design from real-world scenarios.

\item We propose SolutionRAG, which can boost complex engineering solution design through tree-based exploration and bi-point thinking.

\item We conduct extensive experiments, and results show existing methods perform poorly and SolutionRAG is an advanced improvement.
\end{itemize}

% We propose RAG-Star that leverages external
% retrieval to enahnce the deliberative reasoning of
% LLMs based on their internal knowledge.
% • We design an effective retrieval-augmented
% verification and refinement to evaluate and correct
% the inherent reasoning process.
% • We conduct extensive experiments on several datasets, where RAG-Star significantly outperforms existing RAG and reasoning methods.
% \section{SolutionBench for Evaluating Complex Engineering Solution Design}
\section{SolutionBench}

\label{sec:bench}

% \begin{figure}[t!]
% \centering
% \includegraphics[width=\linewidth]{figures/bench_fig.pdf} 
% \caption{Illustration of the SolutionBench construction method, which include collecting technology reports from engineering journals to ensure authority and authenticity, extracting useful content based on a manual generated template and powerful LLMs, harvesting the benchmark after manual checking and merging.}  
% \label{fig:bench}
% \end{figure}

\begin{figure*}[t!]
\centering
\includegraphics[width=\linewidth]{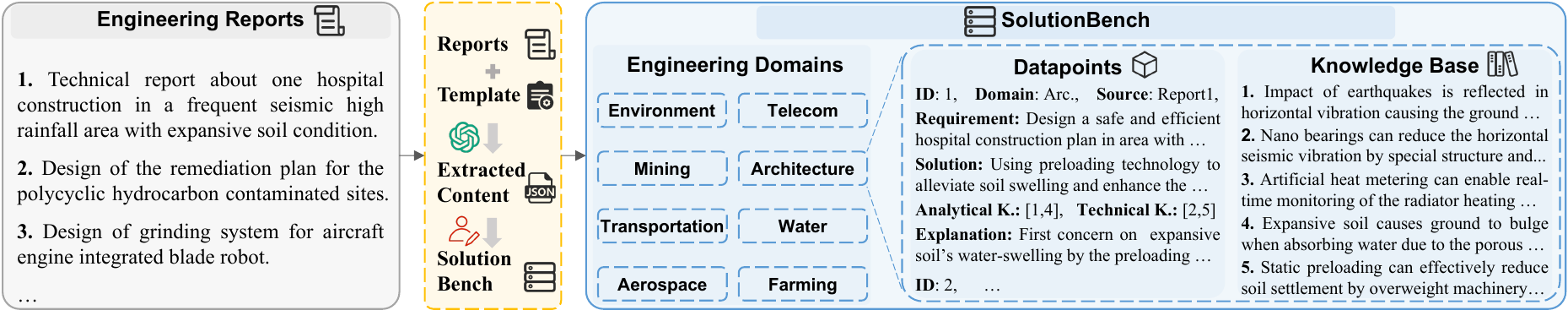} 
\caption{Illustration of the SolutionBench construction method, which includes collecting technology reports from engineering journals to ensure authority and authenticity, extracting useful content based on a manually formatted template and powerful LLMs, and finally harvesting the benchmark after manual verification and merging.}  
\label{fig:bench}
\end{figure*}

% 如上所述，基于RAG技术研究Complex Engineering Solution Design 任务对于提升人类社会的生产效率具有重要价值，然而，此前RAG领域的工作并没有对此进行深入探索。因此，this paper firstly 提出一个新的benchmark，SolutionBench，which 能够用来评估一个系统Designing solutions for complex engineering requirements 的能力。  Specifically, as illustrated in Figure 1，我们首先从各个工程领域的权威期刊中抽取针对复杂需求的方案设计的技术报告，然后，基于人工设计的抽取模版利用GPT进行有效内容的抽取，最后，经过仔细的人工审查和修正后将抽取出的内容整合为完整的RAG benchmark. 下面是构建SolutionBench的详细过程。
As mentioned above, research on complex engineering solution design tasks has significant value in enhancing the productivity of human society, but previous works in RAG field do not explore this in depth. Therefore, this paper introduces a new benchmark, SolutionBench, which can evaluate a system's ability to design solutions for complex engineering requirements. Specifically, as illustrated in Figure~\ref{fig:bench}, we first collect engineering technical reports about complex solution design from authoritative journals across various engineering fields. Then, based on manually formatted extraction templates, we use powerful LLMs to implement useful content extraction. Finally, after  manually checking and removing redundancy, the extracted content is integrated into a complete benchmark. Here is  detailed process of constructing SolutionBench:

\subsection{Authoritative Data Source}
% 在确定benchmark构建的数据源时，我们主要考虑数据权威和真实性以及领域多样性这两个方面。
To ensure the credibility of benchmark, we primarily consider two key factors when determining  data sources: the authority and authenticity of data, as well as the diversity of engineering domains.

\paragraph{Authority and Authenticity.} 
In order to ensure the benchmark's evaluation results can accurately reflect the system's capabilities under real engineering requirements, it is essential to ensure the data sources come from authoritative experts and real-world scenarios. To this end, we select authoritative journals in engineering fields as data sources, choosing engineering reports that involve complex engineering solution design. The requirements in these reports are derived from real industrial scenarios and provided by industry experts under strict peer review, thus ensuring the authenticity and authority of  data sources. The detailed list of used engineering journals is shown in Appendix~\ref{app:journals}.
% 第二，Complex Engineering Solution Design 在各个工程领域中都是迫切的需求，因此构建benchmark的数据源必须涵盖广泛的领域，从而保证能可靠地评测各个系统在广泛的工程领域上的通用性能。To this end，我们基于检索网站上的学科分类机制得到了8个大类：环境工程、矿业、交通运输、航空航天、电力工程、建筑工程、水利工程、农业，从而保证了数据源涵盖了广泛多样的工程领域。
\paragraph{Domain Diversity.} 
Since the need for complex engineering solution design is urgent in multiple engineering domains, the data sources used to construct benchmark must cover a broad range of domains to ensure comprehensive evaluation. To this end, we select authoritative journals from eight major categories based on the discipline classification mechanism of the search websites: Environment, Mining, Transportation, Aerospace, Telecom, Architecture, Water Resource, and Farming. The coverage of these fields ensures that the data sources include diverse engineering scenarios, providing a broad reference for system evaluation.

% \subsection{Template and Extraction}
\subsection{Template-based Extraction via LLM}
% 为了将原始的技术报告转变成方便对系统进行评测和打分的数据格式，我们参考了最先进的generative information extraction技术，使用LLM基于人工设计的template进行content extraction，从每一个技术报告中获取：requirement, solution, analytical knowledge， technical knowledge, and explanation.
To transform original engineering technical reports into data for evaluation and scoring, we format a template manually and extract  following content from each report via LLMs: requirement, solution, analytical knowledge, technical knowledge, and explanation, based on the generative information extraction~\cite{lu2022unifiedstructuregenerationuniversal,zhang-etal-2025-survey}.
% First, 为了便于对系统进行complex engineering solution design 任务的能力进行测试和评分，我们设计的content extraction template包括以下关键字段：（1）Requirement，which is技术报告中所解决的来自于真实工程场景下的复杂需求，（2）Solution，which is 行业顶尖专家设计出的完整可靠的solution，（3）Analytical Knowledge，which is 专家在设计solution过程中对复杂需求做分析时所用到的专业知识（4）Technical Knowledge，which is 专家在应对复杂需求并形成完整solution时所使用的技术的专业知识（5）Explanation，which is 专家时如何利用Analytical Knowledge和Technical Knowledge对复杂需求进行分析并逐渐形成最终的完整solution的，这个explanation可以用来做评估时的一个辅助参考. 在模版的每一个关键字段旁边我们都写清楚该字段的含义，从而使LLM做抽取时明白每一个字段的作，完整的模版如app 所示.
\paragraph{Template.} In order to facilitate the testing and scoring, we format an extraction template including  following keys: (1) Requirement, which refers to the complex needs from real engineering scenarios addressed in reports, (2) Solution, which is the complete and reliable solution designed by top industry experts, (3) Analytical Knowledge, which is the professional knowledge used by experts when analyzing the complex requirements during solution design process (e.g., \textit{Impact of earthquakes is mainly reflected in horizontal vibration}), (4) Technical Knowledge, which is the professional knowledge used by experts to address the complex requirements and develop the complete solutions (e.g., \textit{Nano bearings can reduce the horizontal seismic vibration by  special structure}), (5) Explanation, which outlines how the experts use analytical knowledge and technical knowledge to analyze the complex requirements and gradually design complete solutions. This explanation can serve as an auxiliary reference during the evaluation process.  The complete template used to implement the extraction process is shown in the Appendix~\ref{app:template}.

% Second, 由于原始的技术报告是pdf格式的文件，无法直接进行content extraction，我们首先利用marker将pdf文件转为纯文本，然后将该文本与人工设计的template一起输入给GPT-4o，从而获取抽取后的内容，并将其转变为json格式保存下啦。
\paragraph{Extraction Process.}
Since the original engineering reports are in PDF format and cannot be directly processed for content extraction, we first use the marker tool\footnote{\url{https://github.com/VikParuchuri/marker}} to convert the PDF files into plain text. And then we input the plain text along with the manually formatted template into GPT-4o~\cite{openai2024gpt4ocard},  extracting content as described in the template. Finally we transform extracted content  into JSON format and save it for further process.

\subsection{Manual Data Verification} 
% 为了进一步确保benchmark的可靠性，我们进行了两个步骤的人工审核和修正，从而去除大模型抽取内容中的错误和重复，并最终将所有数据重新整理为RAG格式的benchmark。
To further ensure the credibility of the benchmark, we manually check correctness and remove the redundancy for the extracted content.
% Firstly，由于LLM是一个概率模型，不能保证每一个抽取出来的内容都符合我们的设定，因此，我们人工检查LLM抽取出来的每一个内容，一方面考察该内容是否与原始技术报告中的内容相一致，另一方面考察该内容是否符合模版中的定义，如果遇到不正确的内容，我们直接人工将其修正。
\paragraph{Correctness Checking.} Since the LLM is a probabilistic model and cannot guarantee that every extracted piece of content aligns with our specifications, we manually check each extracted content. On one hand, we examine whether the content matches the information in original engineering reports, on the other hand, we assess whether the content adheres to definitions in the template. For incorrect content, we directly correct it manually.
% Secondly，由于我们在每一个工程领域中都会选择很多技术报告作为数据源，而相同领域中的技术报告中解决实际需求时用到的Analytical Knowledge以及Technical Knowledge可能是相似的甚至是重复的，那么在合成大知识库是就会出现冗余。因此，我们会人工检查同一个领域下不同技术报告构造出的数据中的knowledge是否有重复，如果有重复，我们人工将重复的知识进行合并。
\paragraph{Redundancy Removing.} Since we select many technical reports as data sources for each engineering domain, the analytical knowledge and technical knowledge used to address complex requirements from the same domain may be similar or even identical, resulting in redundancy when constructing a large knowledge base. Therefore, we manually check  duplicates for the knowledge in each domain. If duplicates are found, we manually merge the redundant knowledge to one knowledge.
% Finally，我们将同一个领域下的所有数据的知识合并为一个大的corpus，从而获得RAG格式的benchmark，共包含8个领域，每一个领域都包含以下数据

% \begin{table*}[t!]
% \centering
% \resizebox{\linewidth}{!}{
% \begin{tabular}{lcccccccc}
% \toprule

% \textbf{Item} & \textbf{Enviroment} & \textbf{Mining} & \textbf{Transportation} & \textbf{Aerospace} & \textbf{Telecom} & \textbf{Architecture} & \textbf{Water} & \textbf{Farming} \\

% \midrule

% Dataset & 119 & 117 & 124 & 115 & 116 & 118 & 119 & 122 \\

% \midrule

% Knowledge Base & 554 & 543 & 870 & 802 & 840 & 858 & 802 & 868 \\

% \bottomrule
% \end{tabular}%
% }
% \caption{Statistics of the SolutionBench, which include data and knowledge across 8 engineering domains. The number of datapoint in dataset and number of knowledge in knowledge base are shown above.}
% \label{tab:bench_statics}%
% \end{table*}%

\begin{table}[t]
\centering
\resizebox{0.95\linewidth}{!}{
\begin{tabular}{lcc}
\toprule

\textbf{Engineering Domain} & \textbf{\# Datapoint} & \textbf{\# Knowledge} \\
\midrule
Environment (Env.) & 119 & 554 \\
Mining (Min.)& 117 & 543 \\
Transportation (Tra.)& 124 & 870 \\
Aerospace (Aer.)& 115 & 802 \\
Telecom (Tel.)& 116 & 840 \\
Architecture (Arc.)& 118 & 858 \\
Water Resource (Wat.)& 119 & 802 \\
Farming (Far.)& 122 & 868 \\

 \bottomrule
\end{tabular}%
}
\caption{Statistics of the SolutionBench, which include data and knowledge across eight engineering domains. The number of datapoints in dataset and the number of knowledge in knowledge base are shown above.}
\label{tab:bench_statics}%
\end{table}%

\subsection{Datapoint and Knowledge Base} 

After above manual verification, we do content integrate and get 8 high-quality datasets for the 8 domains,  correspondingly with 8 knowledge base. The detailed statistics of benchmark is in Table~\ref{tab:bench_statics}.

\paragraph{Datapoint Format.} The content of datapoints of every  domain is as following formula:
\begin{equation}
\mathcal{D} = \{q_{i}, s_{i}, \{k^{(a)}_{j}\}_{j=1}^{A_i}, \{k^{(t)}_{j}\}_{j=1}^{T_i}, e_{i}\}_{i=1}^N
\end{equation} where $\mathcal{D}$ is the dataset for one domain, $N$ is data number, $q_i$ is one requirement, $s_i$ is the goldn solution, $k^{(a)}_{j}$ is an analytical knowledge used for $q_i$ and $A_i$ is the total number,  $k^{(t)}_{j}$ is a technical knowledge used for $q_i$ and $T_i$ is the total number.

\paragraph{Knowledge Base.} In order to get the referential knowledge base for each engineering domain, we collect all the $k^{(a)}_{j}$ and $k^{(t)}_{j}$ within the same domain into a large corpus, as shown in following:
\begin{equation}
\mathcal{K} = \cup [\{k^{(a)}_{j}\}_{j=1}^{A_i}, \{k^{(t)}_{j}\}_{j=1}^{T_i}] =\{k_{i}\}_{i=1}^M 
\end{equation} 
where $\mathcal{K}$ is the knowledge base for one domain, and $M$ is the number of knowledge in $\mathcal{K}$.

\paragraph{Evaluation Formulating.}
There are two ways to using SolutionBench for evaluation. The first one is that given a requirement $q$ and expect an reliable solution $\hat{s}$, as shown in following formula:
\begin{equation}
\hat{s}=\mathcal{F}(q) 
\end{equation} 

And the second one is RAG setting, which extra provides the relevant knowledge base for retrieval and augmentation, as shown in following formula:
\begin{equation}
\hat{s}=\mathcal{F}(q, \mathcal{K}) 
\end{equation} 

Since the completion of above tasks requires various engineering expertise, which is prone to hallucination issues in regular-sized LLMs~\cite{jiang-etal-2023-active}, we mainly focus on the RAG setting in this paper. At the same time, we also test some powerful deep reasoning LLMs in  experiments without using RAG, the details are in Section~\ref{sec:experi}.
\begin{figure*}[t!]
\centering
\includegraphics[width=\linewidth]{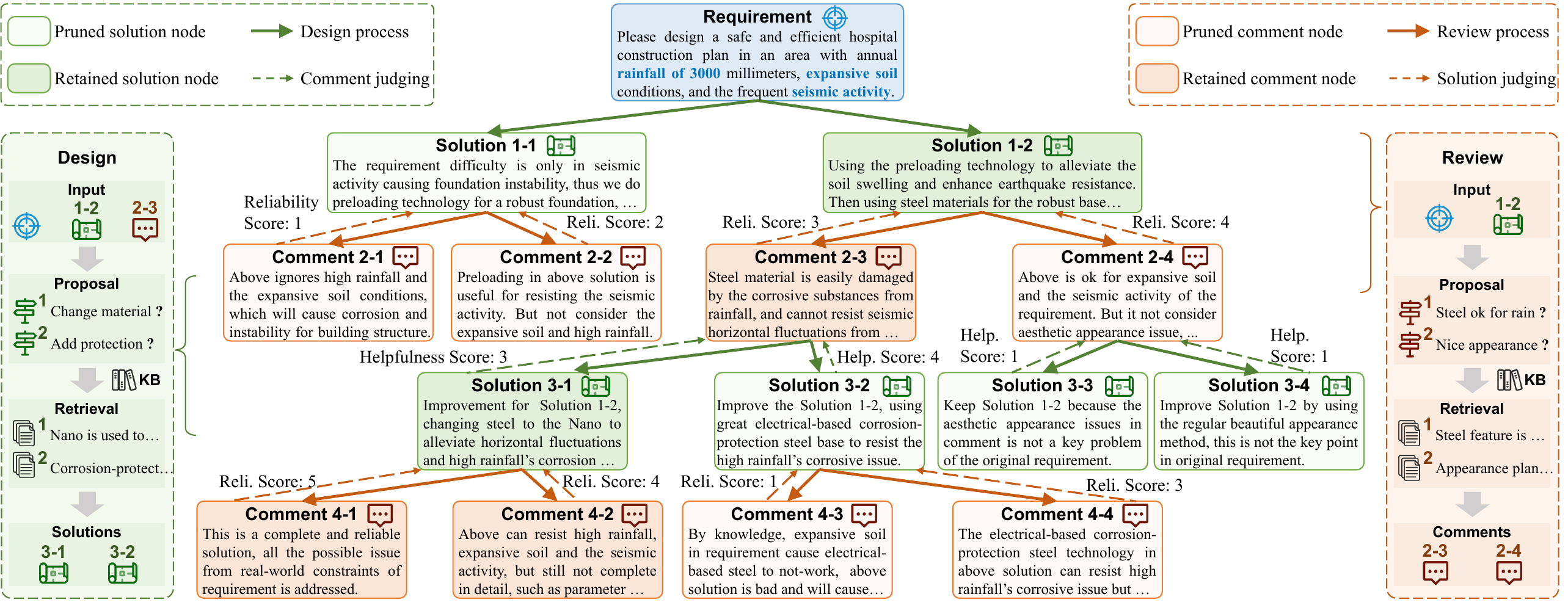} 
\caption{Illustration of SolutionRAG, we set the child number of each node as 2 for easy presentation above. SolutionRAG uses tree-based exploration to find optimal solution improvement process, bi-point thinking to guarantee generated solutions satisfy all constraints, and a pruning mechanism to balance efficiency and performance.}
\label{fig:method}
\end{figure*}

% \section{SolutionRAG via Extensive Exploration and Bidirectional Thinking}
\section{SolutionRAG}

% To further boost complex engineering solution design, we propose SolutionRAG, which generates reliable solutions through bidirectional thinking and an extensive exploration mechanism.

To further advance research in complex engineering solution design, we propose SolutionRAG, a system that can generate reliable solutions through tree-based exploration and bi-point thinking.
Specially, as illustrated in Figure~\ref{fig:method}, since the improvement process from a suboptimal solution to a reliable one is flexible and lacks a fixed reasoning pattern, SolutionRAG performs tree-base exploration to find the most effective improvement process for each input requirement.
Moreover, due to the multiple real-world constraints within the requirements, the system cannot directly guarantee the generated solutions satisfy all constraints. Therefore, SolutionRAG employs a bi-point thinking approach, alternating between solution design and review, gradually enhancing the solution’s completeness and reliability.
Finally, to balance inference performance and efficiency, SolutionRAG employs node evaluation to prune the tree, ensuring that the inference process follows the most promising solutions and the most helpful reviewed comments.

\subsection{Bi-point Thinking Tree}
% In order to 在inference过程中探索出对solution的最佳优化路径，并且使得设计出的solution能够满足requirements中的全部条件，SolutionRAG基于一个Bi-point thinking tree 进行推理，which 由交替连接的solution nodes 以及 comment nodes组成.
To explore the optimal process for  solution improvement during  inference and ensure the output solutions meet all constraints in the requirements, SolutionRAG performs inference based on a bi-point thinking tree, which consists of alternating connected solution nodes and comment nodes.

\paragraph{Solution Node.}
% Solution node 中的内容是针对给定requirement而设计出的solution，which 被期望能够满足requirement中的全部限制条件. 
The content within a solution nodes is the solution designed for the given requirement, which is expected to be a complete and feasible solution meeting all constraints specified in the requirement.
% 在SolutionRAG中，通常树中较浅层的solution node对于requirement的完成度较低，而较深层的完成度较高
The solution nodes at the shallower levels of the tree typically have a lower degree of reliability for the given requirement, while those at deeper levels have a higher degree of reliability.
For convenience, we use $s_j^{(i)}$ represents the $j$-th solution node at the $i$-th layer of the tree.

\paragraph{Comment Node.}
% Comment node中的内容是对某一个solution经过review得到的comment，which 说明了该solution对于给定requirement在哪些方面仍有缺陷. 
The content within a comment node is the comment obtained from reviewing a particular solution, which indicates the aspects in which the solution still has deficiencies with respect to the given requirement.
For convenience of description, we use $c_j^{(i+1)}$ represents the $j$-th comment node at the $(i+1)$-th layer of the tree.

\paragraph{Tree Structure.}
% 上述solution node和comment node以交替连接的方式组成bi-point thinking tree，也就是solution node的子节点都是comment node，而comment node 的子节点都是solution node，如下面的公式所示
The above-mentioned solution nodes and comment nodes are alternately connected to form a bi-point thinking tree, where the child nodes of a solution node are  comment nodes, and the child nodes of a comment node are solution nodes, as shown in the following formula:
\begin{equation}
 s^{(i)}_j \rightarrow \{c^{(i+1)}_h\}_{h=1}^{H}
\end{equation}
% \begin{equation}
% \forall s^{(i)}_j \rightarrow \{c^{(i+1)}_h\}_{h=(j-1)\cdot H + 1}^{j\cdot H}
% \end{equation}
\begin{equation}
 c^{(i+1)}_j \rightarrow \{s^{(i+2)}_h\}_{h=1}^{H}
\end{equation} where $H$ is the number of child node in tree. The content of the root node of the tree is the requirement $q$, so $i$ is at least one in above formula.

\subsection{Solution Improvement via Tree Growth}
% In this section, 我们将会介绍SolutionRAG具体如何通过上述bi-point thinking tree的生长过程实现solution的持续improvement，包括node expansion and node evaluation.
In this section, we introduce how SolutionRAG specifically achieves  continuous improvement of  solutions through the growth process of aforementioned bi-point thinking tree, including the node expansion and node evaluation process.

\paragraph{Node Expansion.}
% (TODO: competeness score helpful score, pruned node, retained score)
% Bi-point thinking tree 生长过程中有两种node expansion过程，一个是基于solution node做review扩展出comment node，另一个是基于requirement或者comment node做design扩展出solution node.
During the growth process of the bi-point thinking tree, there are two types of node expansion actions, one is based on the requirement or comment node to design new solution nodes, and the other is based on reviewing the solution node to create new comment nodes.  

(1) \textbf{\textit{Design}}. 
% 以给定的requirement和某一个comment作为输出 (如果层数大于1)，design的过程首先会通过随机采样的方式通过大模型产生H个proposal，代表H个思考方向.
Given the requirement $q$ and a specific comment $c_{j}^{(i+1)}$ as input (if $i$ is at least one), the design process generate $H$ proposals $\{p_h\}_{h=1}^H$ through random sampling using a LLM, representing $H$ different directions for designing:
\begin{equation}
\{p_h\}_{h=1}^H=\text{LLM}(q,c_{j}^{(i+1)})  
\end{equation}
% 然后基于每一个p都从知识库K中检索回Q个知识文档K_hat，最后将q，c，K_hat，以及上一层的solution s 拼接作为输入，令LLM输出更完善的新solution

Then, small-scale relevant knowledge $\mathcal{K}_h$ is retrieved from the knowledge base $\mathcal{K}$ for each $p_h$:
\begin{equation}
\mathcal{K}_h=\text{Retrieval}(p_h, \mathcal{K}) =\{k_r\}_{r=1}^R
\end{equation}

Finally, $q$, $c_{j}^{(i+1)}$, $\mathcal{K}_h$, and the history solution  $s_{z}^{(i)}$ are concatenated as input, allowing the LLM to output a more refined new solution:
\begin{equation}
s^{(i+2)}_h=\text{LLM}(q,s_z^{(i)},c_{j}^{(i+1)},\mathcal{K}_h)  
\end{equation} 

Thus, we obtain $H$ new solutions $\{s^{(i+2)}_h\}_{h=1}^H$ refined based on the comment $c_{j}^{(i+1)}$. Note that during the generation of solution nodes in the first layer, there are no previous solutions or comments, so we initialize $s_z^{(i)}$ and $c_{j}^{(i+1)}$ as empty text.

(2) \textbf{\textit{Review}}.
% 与上述过程类似，the review 过程也包括三个步骤，即根据q和s生成p，代表H个不同的review direction，然后基于p检索知识，最后基于q，s，k得到对solution 的comment
Similar to the previous process, the review process also consists of three steps: First, proposals $\{p_h\}_{h=1}^H$ are generated based on $q$ and $s_j^{(i)}$, representing $H$ distinct review directions. Next, knowledge $\mathcal{K}_h$ is retrieved for each $p_h$. Finally, comments $\{c^{(i+1)}_h\}_{h=1}^H$ are generated for $s_j^{(i)}$ based on $q$, $s_j^{(i)}$, and $\mathcal{K}_h$. The maximum depth of the bi-point thinking tree, denoted as $L$, is a hyperparameter. Prompts for this part are  Appendix~\ref{app:expansion}.

\paragraph{Node Evaluation.}
% As described in above node expansion part, the number of nodes becomes enormous as the tree growth, leading to significant time consumption during inference. To this end, 我们在树生长的过程中对每一个节点都基于子节点做打分，也就是在完成某动作之后再根据A判断B是否是一个可接受的solution，以及在在完成某动作之后再根据A判断B是否是一个有帮助的comment
As described in above node expansion part, the number of nodes becomes enormous as the tree grows, leading to significant time consumption during inference. To this end, during tree growth, we do prune by each node score from its child nodes, meaning whether $s_j^{(i)}$ is an reliable solution based on $\{c^{(i+1)}_h\}_{h=1}^{H}$ and whether $c_j^{(i+1)}$ is a helpful comment for solution improvement based on $\{s^{(i+2)}_h\}_{h=1}^{H}$. Specifically, for judging $s_j^{(i)}$, we put $s_j^{(i)}$, $c^{(i+1)}_h$, and a suffix $u_s$ together as the LLM input, and get the \textit{reliability score} $\mathcal{J}_h(s_j^{(i)})$ by calculating LLM-predicted average logits of $u_s$:
\begin{equation}
\mathcal{J}_h(s_j^{(i)})=\text{Logits}(u_s|s_j^{(i)},c^{(i+1)}_h)  
\end{equation} where $u_s$ is ``According to the comment, above solution is reliable''. And then get final score $\mathcal{J}(s_j^{(i)})$ for $s_j^{(i)}$ by average all $\{\mathcal{J}_h(s_j^{(i)})\}_{h=1}^H$. Similarly, for judging $c^{(i+1)}_j$, we use $s^{(i)}_z$, $c^{(i+1)}_j$, $s^{(i+2)}_h$, and $u_c$ as input, and get the \textit{helpfulness score} $\mathcal{J}_h(c_j^{(i+1)})$ by calculating LLM-predicted average logits:
\begin{equation}
\mathcal{J}_h(c_j^{(i+1)})=\text{Logits}(u_c|s_z^{(i)},c^{(i+1)}_j,s_h^{(i+2)})  
\end{equation} 
where $u_c$ is ``Comparing the new solution and old solution, the comment is helpful'', and get $\mathcal{J}(c_j^{(i+1)})$ after same averaging process. 

During the tree growth, for each layer we only keep the $W$ highest-scoring nodes, aiming to keep the inference process always focus on the most promising solutions and the most helpful reviewed comments, thus achieving a balance between efficiency and performance. The nodes that are used in final inference process are called \textit{retained nodes}, while those not-used are \textit{pruned nodes}.

\section{Experiments}
\label{sec:experi}

\begin{table*}[t!]
\centering
\resizebox{\linewidth}{!}{
\begin{tabular}{lcccccccccccccccc}
\toprule

\multirow{2}[1]{*}{\textbf{Method}} & \multicolumn{2}{c}{\textbf{Env.}} & \multicolumn{2}{c}{\textbf{Min.}} & \multicolumn{2}{c}{\textbf{Tra.}} & \multicolumn{2}{c}{\textbf{Aer.}} & \multicolumn{2}{c}{\textbf{Tel.}} & \multicolumn{2}{c}{\textbf{Arc.}} & \multicolumn{2}{c}{\textbf{Wat.}} & \multicolumn{2}{c}{\textbf{Far.}} \\
  
\cmidrule(lr){2-3} \cmidrule(lr){4-5}  \cmidrule(lr){6-7} \cmidrule(lr){8-9} \cmidrule(lr){10-11} \cmidrule(lr){12-13}  \cmidrule(lr){14-15} \cmidrule(lr){16-17}
& AS & TS & AS & TS & AS & TS & AS & TS & AS & TS & AS & TS & AS & TS & AS & TS \\

\midrule
\rowcolor[rgb]{ .906,  .902,  .902} \multicolumn{17}{c}{Deep Reasoning Models} \\
\midrule

o1-2024-12-17~\cite{openai2024openaio1card} & 60.5 & 48.3 & 51.9 & 37.5 & 57.3 & 44.7 & 57.8 & 47.6 & 63.5 & 52.3 & 61.2 & 52.0 & 59.9 & 50.4 & 62.9 & 52.2 \\
GLM-Zero-Preview~\cite{glmzero} & 47.0 & 30.6 & 43.2 & 22.2 & 45.2 & 27.0 & 42.3 & 25.7 & 45.1 & 31.7 & 47.7 & 32.4 & 47.3 & 30.8 & 51.4 & 36.6 \\
QwQ-32B-Preview~\cite{qwq} & 54.3 & 38.7 & 48.0 & 27.9 & 47.2 & 29.3 & 47.4 & 31.9 & 52.2 & 35.9 & 51.3 & 35.6 & 49.2 & 33.0 & 53.4 & 37.0 \\

\midrule
\rowcolor[rgb]{ .906,  .902,  .902} \multicolumn{17}{c}{Single-round RAG Methods} \\
\midrule

Naïve-RAG~\cite{NEURIPS2020_6b493230} & 64.8 & 62.2 & 57.2 & 40.1 & 62.7 & 54.9 & 67.7 & 65.4 & 67.4 & 66.8 & 66.2 & 63.3 & \textbf{66.0} & 57.5 & 65.7 & 63.0 \\
Rerank-RAG~\cite{li2023generaltextembeddingsmultistage} & 62.7 & 60.7 & 53.4 & 38.4 & 60.0 & 49.7 & 65.6 & 65.2 & 66.1 & 63.4 & 66.4 & 62.8 & 64.1 & 55.4 & 64.0 & 59.7 \\

\midrule
\rowcolor[rgb]{ .906,  .902,  .902} \multicolumn{17}{c}{Multi-round RAG Methods} \\
\midrule

Self-RAG~\cite{asai2024selfrag} & 64.2 & 63.6 & 56.1 & 41.6 & 62.9 & 56.5 & 68.8 & 69.9 & 67.6 & 66.9 & 66.7 & 65.9 & 64.8 & 58.6 & 65.1 & 61.1 \\
GenGround~\cite{shi-etal-2024-generate} & 54.8 & 46.1 & 53.0 & 33.3 & 54.7 & 37.2 & 55.7 & 46.0 & 58.3 & 50.7 & 60.1 & 50.7 & 60.4 & 48.9 & 59.8 & 52.7 \\
RQ-RAG~\cite{chan2024rqrag} & 53.5 & 44.4 & 48.9 & 28.7 & 53.8 & 38.8 & 55.0 & 46.1 & 57.9 & 44.6 & 56.3 & 46.9 & 54.3 & 39.8 & 57.2 & 45.2 \\

\midrule
\rowcolor[rgb]{ .906,  .902,  .902} \multicolumn{17}{c}{Tree-based Exploration and Bi-point Thinking} \\
\midrule

SolutionRAG (Ours) & \textbf{66.4} & \textbf{67.9} & \textbf{59.7} & \textbf{50.5} & \textbf{64.1} & \textbf{58.5} & \textbf{69.9} & \textbf{72.7} & \textbf{68.8} & \textbf{69.0} & \textbf{67.9} & \textbf{68.0} & \textbf{66.0} & \textbf{60.7} & \textbf{66.9} & \textbf{65.2} \\

\bottomrule
\end{tabular}%
}
\caption{Main experimental results on SolutionBench with eight engineering domains, the AS is the analytical score and TS is the technical score. The table shows that previous methods perform poorly for complex engineering solution design. In contrast, our SolutionRAG is able to output more complete and reliable solutions.}
\label{tab:main_results}%
\end{table*}%

\subsection{Experimental Settings}

% 由于SolutionBench中期待的系统输出是一个可能有各种文字表达形式的solution，规则式metric很难给出符合人类习惯的评分。因此，我们借鉴了此前一些Long-form QA工作的评估方式，使用LLMs作为打分器，计算两个分数，（1）Analytical Score. 将专家产生的solution、专家用到的Analytical knowledge、以及专家形成solution的explanation整合在一起作为reference，让LLMs判断系统输出的solution是否能够像专家产生的solution一样，使用正确的Analytical knowledge对requirements中的复杂限条件进行了充分的考量。（2）Technical Score. 类似地，将专家产生的solution、专家用到的Technical knowledge、以及专家形成solution的explanation整合在一起作为reference，让LLMs判断系统输出的solution是否能够像专家产生的solution一样，使用正确的Technical knowledge来应对requirements中的复杂限条件。
\paragraph{Evaluation Metrics.} Since expected system output in SolutionBench are solutions that may have various textual expressions, rule-based metrics are difficult to provide a score that aligns with human habits~\cite{xu-etal-2023-critical,10.1145/3626772.3657846}. To this end, we follow metrics of previous Long-form QA evaluation methods~\cite{tan2024proxyqa,wang2024leavedocumentbehindbenchmarking}, using GPT-4o\footnote{\url{https://openai.com/index/hello-gpt-4o/}} as score evaluator to compute two scores, (1) \textbf{Analytical Score}: We integrate the expert-designed solution, analytical knowledge used by experts, and the explanation as  reference, and then let evaluator judge whether the system-designed solution, like the expert-designed one, uses the correct analytical knowledge to adequately analysis the complex constraints in requirements. (2) \textbf{Technical Score}: Similarly, we integrate the expert-designed solution,  technical knowledge used by experts, and the explanation  as reference, and then let evaluator judge whether the system-designed solution, like the expert-designed one, uses the correct technical knowledge to tackle the complex constraints in requirements. Both analytical score and technical score are range from 0 to 100. Prompts for this part are in Appendix ~\ref{app:judge}.

% 为了全面地考察此前各个类型的系统解决Complex Engineering Solution Design tasks 的能力，我们广泛地选取了多种类型的mehtos作为baseline. 具体来说，（1）Deep Reasoning Models: 包括ABCD，这些模型具备强大的长思维链推理能力，但是并没有像RAG一样使用外部知识（2）Single-round RAG Methods: 这些方法只进行一轮检索和生成，其中naiive rag对检索结果没有任何处理，而Rerank-RAG会利用一个额外的模型对检索结果进行重排（3）Multi-round RAG Methods：这些方法会进行多轮RAG，即迭代地进行 问题改写，检索，过滤和利用，经过不断迭代，能够逐渐提高检索准确率
\paragraph{Selected Baselines.}
In order to comprehensively evaluate the abilities of various types of systems in solving complex engineering solution design tasks, we extensively select multiple types of methods as baselines in the experiments. Specifically, (1) \textbf{Deep Reasoning Models}: This type includes models like o1-2024-12-17~\cite{openai2024openaio1card}, GLM-Zero-Preview~\cite{glmzero}, and QwQ-32B-Preview~\cite{qwq}, which possess strong long-chain reasoning capabilities, but do not utilize external knowledge like RAG. (2) \textbf{Single-round RAG Methods}: These methods perform only one round of retrieval and generation, where Naive-RAG~\cite{NEURIPS2020_6b493230} does not process the retrieval results, while Rerank-RAG~\cite{li2023generaltextembeddingsmultistage} uses an additional model to re-rank the retrieval results. (3) \textbf{Multi-round RAG Methods}: These methods conduct multiple rounds of RAG, iteratively performing tasks such as question rewriting, retrieval, filtering, and generating intermediate answer. We choose 3 accredited methods, which are Self-RAG~\cite{asai2024selfrag}, GenGround~\cite{shi-etal-2024-generate}, and RQ-RAG~\cite{chan2024rqrag}.

\begin{table*}[t!]
\centering
\resizebox{\linewidth}{!}{
\begin{tabular}{lcccccccccccccccccc}
\toprule

\multirow{2}[1]{*}{\textbf{Method}} & \multicolumn{2}{c}{\textbf{Env.}} & \multicolumn{2}{c}{\textbf{Min.}} & \multicolumn{2}{c}{\textbf{Tra.}} & \multicolumn{2}{c}{\textbf{Aer.}} & \multicolumn{2}{c}{\textbf{Tel.}} & \multicolumn{2}{c}{\textbf{Arc.}} & \multicolumn{2}{c}{\textbf{Wat.}} & \multicolumn{2}{c}{\textbf{Far.}} & \multicolumn{2}{c}{\textbf{Overall}} \\
  
\cmidrule(lr){2-3} \cmidrule(lr){4-5}  \cmidrule(lr){6-7} \cmidrule(lr){8-9} \cmidrule(lr){10-11} \cmidrule(lr){12-13}  \cmidrule(lr){14-15} \cmidrule(lr){16-17} \cmidrule(lr){18-19}
& AS & TS & AS & TS & AS & TS & AS & TS & AS & TS & AS & TS & AS & TS & AS & TS & AS & TS \\

\midrule

SolutionRAG & \textbf{66.4} & \textbf{67.9} & \textbf{59.7} & \textbf{50.5} & \textbf{64.1} & \textbf{58.5} & \textbf{69.9} & \textbf{72.7} & \textbf{68.8} & \textbf{69.0} & \textbf{67.9} & \textbf{68.0} & \textbf{66.0} & \textbf{60.7} & \textbf{66.9} & \textbf{65.2} & \textbf{66.2} & \textbf{64.1} \\

~~w/o tree structure & 63.5 & 66.5 & 57.3 & 46.2 & 63.1 & 57.4 & 60.8 & 68.4 & 60.9 & 63.7 & 66.2 & 67.2 & 65.6 & 59.9 & 64.2 & 63.9 & 62.7 & 61.7 \\

~~w/o bi-point thinking & 62.8 & 64.7 & 55.6 & 47.3 & 61.5 & 55.7 & 63.2 & 68.3 & 62.6 & 64.8 & 67.5 & 67.3 & 64.4 & 59.1 & 65.2 & 64.7 & 62.9 & 61.5 \\

\bottomrule
\end{tabular}%
}
\caption{Ablation results for tree-based exploration and bi-point thinking. The table shows that both mechanisms have obviously positive effects for SolutionRAG and exhibit a similar level of importance in the overall.}
\label{tab:ablation_results}
\end{table*}

% 对于baseline中的Deep Reasoning Models，我们直接使用官方提供的API进行实验。对于SolutionRAG以及baselines中的所有RAG方法，为了保证对比的公平性，我们都使用相同的实施方案：基座模型是Qwen2.5-7B-Instruct，检索模型是NV-Embed-v2，检索结果保留Top-10. 对于SolutionRAG，我们设定树的最大深度为5，每个节点扩展两个子节点，剪枝时每一层只保留一个得分最高的节点。为了便于实验，我们在进行所有基于RAG的实验时，都会将base model通过vllm部署为api，
\paragraph{Implementation Details.}
For deep reasoning models in baselines, we directly use official API for experiments. For the single-round and multi-round RAG methods, we follow their official  process. For SolutionRAG, we set maximum tree depth $L$ as 5, number of child per node $H$ as 2, and number of retained node $W$ in pruning as 1. To ensure fair comparison, we adopt the following same implementation setting for SolutionRAG and all RAG-based methods in baselines: base model is Qwen2.5-7B-Instruct~\cite{qwen2.5}, retrieval model is NV-Embed-v2~\cite{lee2025nvembedimprovedtechniquestraining}, and the number of retrieval results $R$ is setting as 10. For convenience, in all RAG-based experiments, we deploy the base model as API by vLLM\footnote{\url{https://pypi.org/project/vllm/}}.

\subsection{Overall Results}

Results compared with baselines are shown in the Table~\ref{tab:main_results}, there are two main conclusions:

% （1）此前的方法无法有效解决complex engineering solution design任务. The table shows that，一方面，没有使用RAG的deep reasoning models在SolutionBench的全部8个领域的测试中都只能取得比较低的分数，例如GLM-Zero-Preview在aerospace领域上的analytical score只有42.3。 另一方面，使用RAG的methods能够取得稍高一些的性能，但是仍然处于比较低的水平，例如Naive-RAG在mining领域的technical score只有40.1，Self-RAG在environmental engineering领域上的technical score只有63.6，这些实验结果说明complex engineering solution design任务对于此前的方法是一个挑战
\textbf{Previous methods fail to effectively address the complex engineering solution design.}
The table shows that, on one hand, deep reasoning models without RAG perform poorly across all eight domains in SolutionBench. For example, GLM-Zero-Preview achieves an analytical score of only 42.3 in the aerospace domain.
On the other hand, RAG-based methods achieve some better performance but still remain at relatively low levels. For instance, Naive-RAG obtains a technical score of only 40.1 in the mining engineering  domain, and Self-RAG achieves a technical score of just 63.6 in the environmental engineering domain.

% （2）相比之下，SolutionRAG是address complex engineering solution design任务的一个有效的系统. The table shows that SolutionRAG在benchmark上的所有领域的评测中都取得了SOTA的性能，相对于baselines展现出明显的性能提升。例如，在mining领域的technical score上，SolutionRAG相比于Naiive-RAG提升了10.4，相比于Self-RAG提升了8.9。这些实验结果证明了SolutionRAG能够比较好地完成实际工程场景的复杂方案设计任务，从而在工业生产活动中节省人力和时间资源。
\textbf{In contrast, SolutionRAG is an effective system for complex engineering solution design tasks.}
The table shows that SolutionRAG achieves SOTA performance across all of eight domains in the benchmark, demonstrating a significant improvement over baseline methods.
For example, in the mining domain, SolutionRAG improves the technical score by 10.4 compared to Naive-RAG and by 8.9 compared to Self-RAG.
These experimental results confirm that SolutionRAG can effectively handle complex solution design tasks in various real-world engineering scenarios.

% \subsection{Ablation Results of Extensive Exploration and Bidirectional Thinking}
% Extensive exploration and bidirectional thinking是SolutionRAG中的两个关键机制，为了探究这两个机制的作用，我们对其进行消融实验，results are shown in Table2， there are two main conclusions
\subsection{Ablation Results}
Since tree-based exploration and bi-point thinking are two key mechanisms in SolutionRAG, we conduct two ablation experiments, results are shown in Table~\ref{tab:ablation_results}, where ``w/o tree structure'' is that each node generates only one child, resulting in a single-chain inference pattern, and ``w/o bi-point thinking'' is that the tree does not include reviewing and all nodes are solutions, leading to a uni-point thinking inference pattern. There are two main conclusions:

% Both extensive exploration and bidirectional thinking 都具有正向作用. The table shows that 去掉任意一个机制之后性能都会明显下降，说明这两个机制确实是解决complex engineering solution design tasks的核心。
\textbf{Both of the tree-based exploration and bi-point thinking have positive effects.} The table shows that removing either mechanism leads to a significant decline in performance, indicating that these two mechanisms are indeed central to solving complex engineering solution design tasks.

% Extensive exploration and bidirectional thinking 呈现出相近的重要程度。The table shows that去掉这两个机制之后，综合来看性能下降幅度想接近，这说明这两个机制在SolutionRAG中占有相近的重要程度
\textbf{Tree-based exploration and bi-point thinking exhibit a similar level of importance}. The table shows that after removing these two mechanisms, overall performance decline is quite similar, indicating these two mechanisms hold a comparable level of importance in SolutionRAG.

\begin{figure}[t!]
\centering
\includegraphics[width=0.9\linewidth]{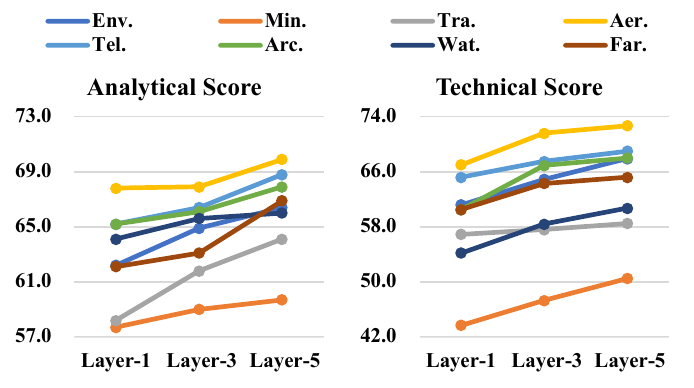} 
\caption{Performance changes during the tree growth. The figure shows that scores become higher as the tree grows, proving SolutionRAG can indeed improve the solution scores as inference being deep.} 
\label{fig:depth}
\end{figure}

\subsection{Detailed Analysis}
In order to further validate the  SolutionRAG, we do some detailed analysis, including performance changeing during the tree growth process and effectiveness of the node evaluation in SolutionRAG.

\textbf{Performance during Tree Growth.}
% 验证SolutionRAG的一个重要方面就是随着树深度的增大，在extensive exploration和bidirectional thinking两个机制的共同作用下，系统产生的solution是否真的在逐渐变得更好. 为了探究这一点, 我们取树中的shallow层的solution、middle层的solution以及deep层的solution进行评分，实验结果如Table 3 所示，从shallow层到Deep层的性能逐渐提升，这证明了SolutionRAG确实能够随着inference time的累积而提升solution的得分.
To examine whether the solutions actually improve as the tree depth increases in SolutionRAG inference, we score the solutions from the layer-1, 3, and 5 of the tree. The experimental results are shown in Figure~\ref{fig:depth}, performance gradually improves from the shallow layer to the deep layer, which proves that \textit{SolutionRAG can indeed improve the solution as inference process being deep}.

\begin{figure}[t!]
\centering
\includegraphics[width=0.9\linewidth]{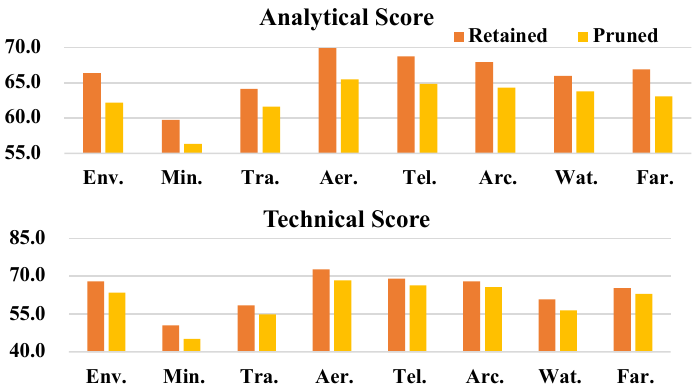} 
\caption{Effectiveness of node evaluation mechanism. The figure shows that scores in retained nodes are higher than in pruned nodes, thus the node evaluation is an effective method for judging and pruning in SolutionRAG.} 
\label{fig:unused}
\end{figure}

\textbf{Effectiveness of Node Evaluation.}
To examine whether node evaluation mechanism for pruning the tree is effective, we compare the scores of solutions from the retained nodes with those from the pruned nodes. The results are shown in Figure~\ref{fig:unused}, where the scores of solutions from the retained nodes are significantly higher than pruned nodes, which proves that \textit{node evaluation is an effective mechanism for judging and pruning}.

% \subsubsection{Case Study: Comparing  SolutionRAG and Prior Multi-round RAG}

\section{Related Work}

% RAG 技术能够在解决知识型问题时通过检索和利用必要的外部知识对模型进行辅助，从而缓解LLM的幻觉问题。早期的RAG任务主要集中在普通的single-hop QA任务，随着RAG技术的发展，最近的研究聚焦在通过multi-round RAG解决需要一定推理或者整合的Multi-hop QA任务或者Long-form QA任务.

% RAG can assist models in solving knowledge-based problems by retrieving and utilizing necessary external knowledge, thereby alleviating the hallucination problem of LLMs~\cite{jiang-etal-2023-active,li2025structrag}. Early RAG tasks primarily focused on standard Single-hop QA tasks~\cite{47761,joshi2017triviaqalargescaledistantly}. With the development of RAG technology, recent papers shift towards addressing Multi-hop QA ~\cite{ho-etal-2020-constructing,wu2024cofcastepwisecounterfactualmultihop} or Long-form QA ~\cite{tan2024proxyqa,qi-etal-2024-long2rag} that require a certain level of reasoning or integration through Multi-round RAG methods~\cite{tran2024rareretrievalaugmentedreasoningenhancement,yu2024auto}.

\paragraph{Complex QA Tasks.}
% 此前RAG领域的工作主要关注需要一定推理的知识型问答任务，包括Multi-hop QA以及Long-form QA.
Recent works in the RAG field mainly focused on knowledge-based question answering tasks that require some level of reasoning.
(1) \textbf{Multi-hop QA.} The question  is a combination of multiple sub-questions, and the expected answer is an entity fragment from relevant knowledge documents~\cite{yang2018hotpotqadatasetdiverseexplainable,ho-etal-2020-constructing,zhu2024fanoutqa,wu2024cofcastepwisecounterfactualmultihop}. 
(2) \textbf{Long-form QA.} The question is an open-ended and comprehensive question, and the expected answer is a text paragraph formed by integrating knowledge fragments from relevant documents~\cite{fan-etal-2019-eli5,stelmakh-etal-2022-asqa,tan2024proxyqa,qi-etal-2024-long2rag}. 
Compared to above two tasks, questions of complex engineering solution design are with multiple real-world constraints. And the expected answer is a solution needing  flexible improvement process, rather than  an entity fragment or simply integrated paragraph. Therefore, complex engineering solution design is a novel and challenging task.

\paragraph{Advanced RAG.}
Prior advanced RAG systems use a multi-round approach to iteratively perform  rewriting, retrieval, reranking, and generating intermediate answers~\cite{asai2024selfrag,shi-etal-2024-generate,chan2024rqrag,wang2024retrieverandmemoryadaptivenoteenhancedretrievalaugmented,tran2024rareretrievalaugmentedreasoningenhancement,yu2024auto}. 
Compared to these systems, SolutionRAG is with a bi-point thinking tree, which can respond to challenges of complex engineering solution design. 
Recently some papers construct RAG systems based on MCTS, achieving better performance through deep thinking~\cite{jiang2024ragstarenhancingdeliberativereasoning,li2025searcho1agenticsearchenhancedlarge,wu2025talkrightspecialistsrouting}. However, these methods lack a mechanism to ensure that all engineering requirements are met, thus failing to guarantee the reliability of solutions.

\section{Conclusion}

In this paper, we first construct SolutionBench based on engineering reports across various domains, which can examine the ability of systems on complex engineering solution design. Further, we propose SolutionRAG, which explore the optimal solution-improvement process and gradually generates reliable solutions  by a bi-point thinking tree. In experiments, previous methods perform poorly in complex engineering solution design task, while SolutionRAG represents a good improvement over existing approaches. This paper offers a promising direction and can inspire the further research.

\section{Limitations}

% Complex engineering solution design 是一个需要系统基于专业知识进行深度推理的任务，需要模型具有强大的问题分析、方案推理以及批判性思考的能力。在this paper中由于GPU计算资源有限，我们并没有对LLMs在这些方面进行训练，而是limit在LLMs自身的已有能力上。未来使用强化学习等方式对LLMs进行训练是获取更强的complex engineering solution design系统的一个可行方向。此外，同样由于GPU计算资源有限，我们在实验中没有对树的宽度和深度等超参数做广泛的探究，这可能是未来工作的一个有价值的研究问题.

Complex engineering solution design is a task requiring deep research based on professional knowledge, which demands  the model has strong capabilities in problem analysis, solution reasoning, and critical thinking. In this paper, due to limited GPU computational resources, we construct the system by the existing capabilities of LLMs, without considering special training. Therefore, a possible direction for future work is to use reinforcement learning to train LLMs, in order to develop more powerful complex engineering solution design systems. Additionally, due to the same limitation in GPU computational resources, we do not extensively explore hyperparameters such as the width and depth of the tree in our experiments. This could be a valuable research topic for future work.
% Bibliography entries for the entire Anthology, followed by custom entries
%\bibliography{anthology,custom}
% Custom bibliography entries only
\bibliography{custom}

\newpage

\appendix

\section{List of Engineering Journals}
\label{app:journals}
In order to ensure that the data sources used to construct the benchmark are authentic, authoritative, and diverse, we select engineering reports on solution design from authoritative journals in multiple engineering fields as our data sources. If the report is in Chinese, we extract the useful content, then use GPT-4o to translate the content into English and manually verify its accuracy. We list the used engineering journals, including the journal name and ISSN meaning the international standard serial number. The detailed list of used engineering journals is shown in  Table~\ref{tab:list1}, ~\ref{tab:list3} and \ref{tab:list2}. 

\section{Template for Extraction}
\label{app:template}
\begin{figure*}[t!]
\centering
\includegraphics[width=\linewidth]{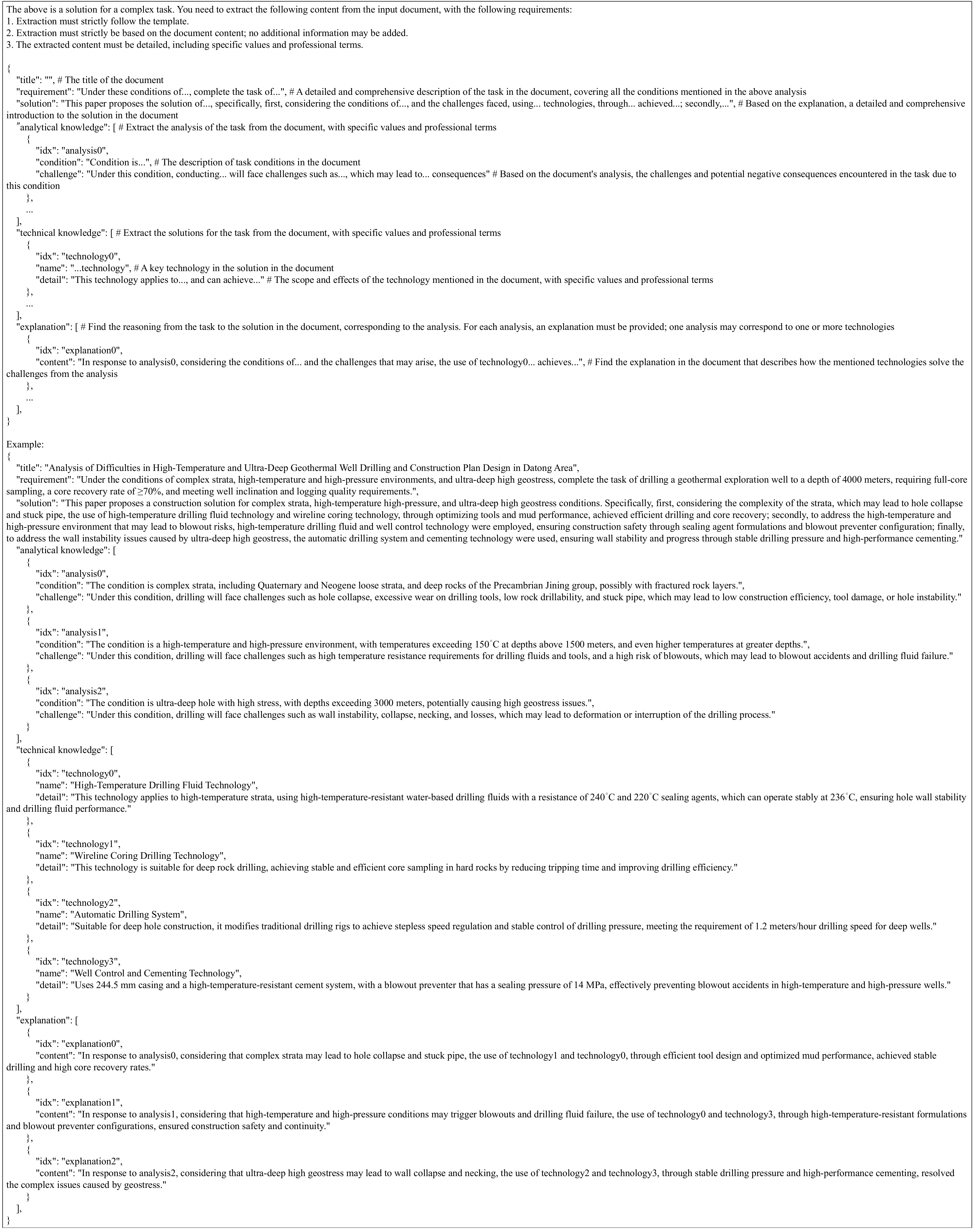} 
\caption{Template used to extract useful content from original engineering reports, aiming to capture real-world complex requirements, expert-authored solutions, analytical knowledge used to interpret the requirements, technical knowledge applied in addressing the requirements, and explanations for the expert's solution design process.}
\label{fig:template}
\end{figure*}

In order to obtain the necessary content for evaluating and judging systems from engineering reports, we manually format a template. When extracting, we combine the report with this template, input it into GPT-4o, and then organize the output into JSON format and save it. The extracted content includes: real-world complex requirements, expert-authored solutions, analytical knowledge used to interpret the requirements, technical knowledge applied in addressing the requirements, and explanations for the expert's solution design process. The  complete template is shown in Figure~\ref{fig:template}.

\section{Prompt for Node Expansion}
\label{app:expansion}
\begin{figure*}[t!]
\centering
\includegraphics[width=\linewidth]{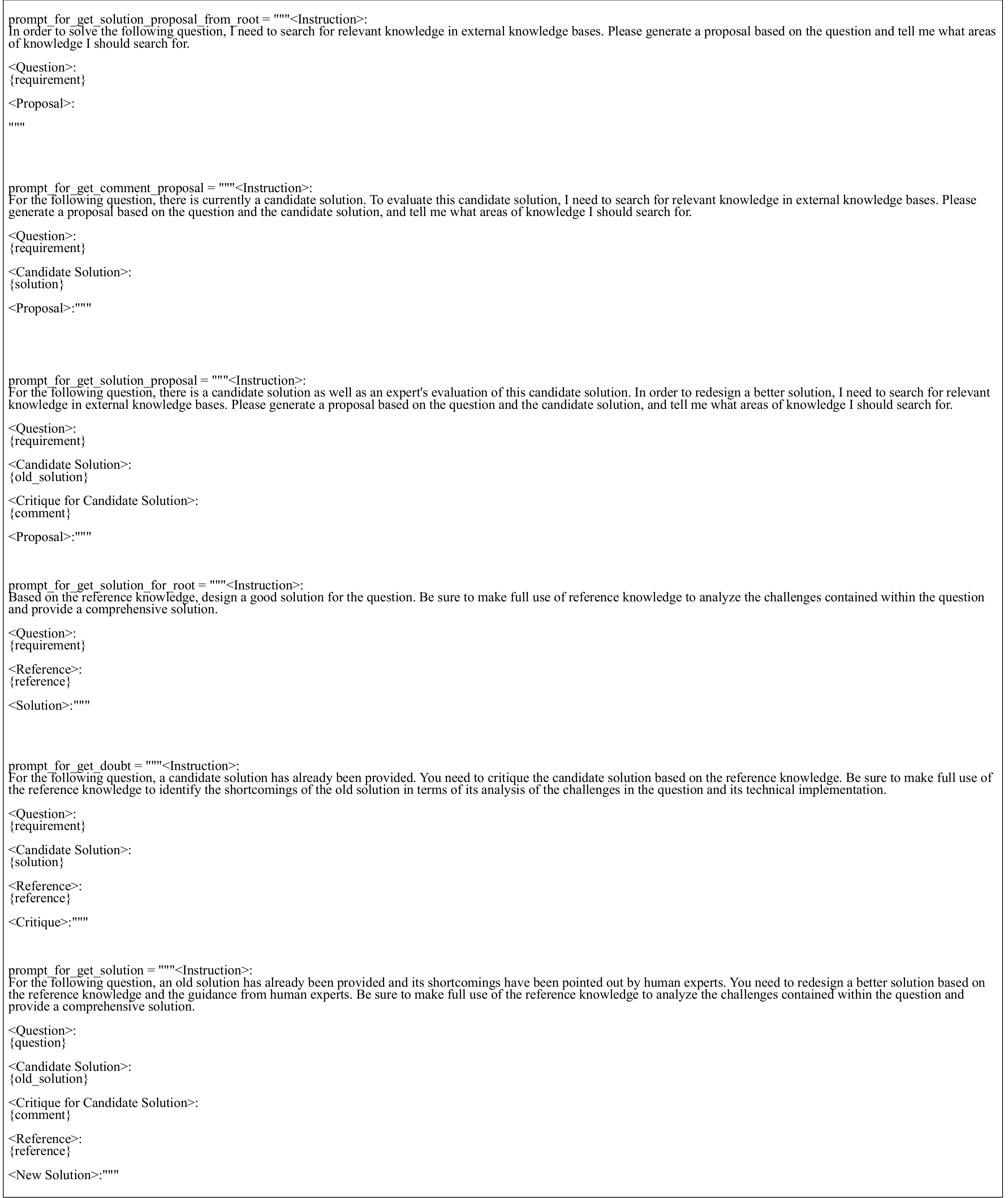} 
\caption{Prompts used in node expansion of tree growth, including generating solution proposals and solutions based on the root node, generating comment proposals and comments based on a solution node, and generating  solution proposals and solutions based on a comment node.}
\label{fig:grow}
\end{figure*}

In the growth of the tree, there are two expansion processes: design and review. The review process is divided into two stages: generating proposals based on parent node information and generating comments based on retrieved documents. The design process is also divided into two stages: generating proposals based on parent node information and generating solutions based on retrieved documents. Moreover, the design process based on the root node and the design process based on the comment node use different prompts due to the differences in input information. All the prompts mentioned in this section are shown in Figure~\ref{fig:grow}.

% \begin{table*}[t!]
% \centering
% \resizebox{\linewidth}{!}{
% \begin{tabular}{lcccccccc}
% \toprule

% \textbf{Item} & \textbf{Enviroment} & \textbf{Mining} & \textbf{Transportation} & \textbf{Aerospace} & \textbf{Telecom} & \textbf{Architecture} & \textbf{Water} & \textbf{Farming} \\

% \midrule

% Dataset & 119 & 117 & 124 & 115 & 116 & 118 & 119 & 122 \\

% \midrule

% Knowledge Base & 554 & 543 & 870 & 802 & 840 & 858 & 802 & 868 \\

% \bottomrule
% \end{tabular}%
% }
% \caption{Statistics of the SolutionBench, which include data and knowledge across 8 engineering domains. The number of datapoint in dataset and number of knowledge in knowledge base are shown above.}
% \label{tab:bench_statics}%
% \end{table*}%

\begin{table}[t]
\footnotesize
\centering
\resizebox{0.98\linewidth}{!}{
\begin{tabular}{lc}
\toprule

\multicolumn{2}{c}{\textbf{Environment}} \\ 
\cmidrule(lr){1-1} \cmidrule(lr){2-2} 
\textbf{Journal Name} & \textbf{ISSN}  \\ 
\midrule

Journal   of Environmental Engineering Technology                    & 1674-991X \\
Environmental Sanitation Engineering                                 & 1005-8206 \\
The Administration and Technique of Environmental Monitoring         & 1006-2009 \\
Environment and Development                                          & 2095-672X \\
Environmental Protection and Technology                              & 1674-0254 \\
Green Environmental Protection Building Materials                    & 1673-6680 \\
Journal of Henan University of Urban Construction                    & 1674-7046 \\
Urban Management and Science \& Technology                           & 1008-2271 \\
Science and Technology Square                                        & 1671-4792 \\
Construction   Materials \& Decoration                               & 1673-0038 \\
Intelligent City                                                     & 2096-1936 \\
Instrument Standardization \& Metrology                              & 1672-5611 \\
Northwest Hydropower                                                 & 1006-2610 \\
Technology \& Economics in Petrochemicals                            & 1674-1099 \\
Water Purification Technology                                        & 1009-0177 \\
Construction Science and Technology                                  & 1671-3915 \\
Urban Geology                                                        & 2097-3764 \\
Engineering   and Construction          & 1673-5781 \\
Engineering and Technological Research  & 2096-2789 \\
Scientific and Technological Innovation & 2096-4390 \\
Engineering \& Test                     & 1674-3407 \\
Inner   Mongolia Water Resources                                     & 1009-0088 \\
China Cement                                                         & 1671-8321 \\
Guangdong Chemical Industry                                          & 1007-1865 \\
Jiangxi Building Materials                                           & 1006-2890 \\
Tianjin Science \& Technology                                        & 1006-8945 \\
Journal of Zhejiang University of Water Resources and Electric Power & 2095-7092 \\
China Municipal Engineering                                          & 1004-4655 \\
China Storage \&   Transport                                         & 1005-0434 \\

\midrule
\midrule
\multicolumn{2}{c}{\textbf{Mining}} \\ 
\cmidrule(lr){1-1} \cmidrule(lr){2-2} 
\textbf{Journal Name} & \textbf{ISSN}  \\ 
\midrule

Coal   Engineering                           & 1671-0959 \\
Mining Engineering                           & 1671-8550 \\
Mechanical Management and Development        & 1003-773X \\
Coal and Chemical Industry                   & 2095-5979 \\
Colliery Mechanical \& Electrical Technology & 1001-0874 \\
Modern Mining                                & 1674-6082 \\
China Mine Engineering                       & 1672-609X \\
Shandong Coal Science and Technology         & 1005-2801 \\
Jiangxi Coal Science \& Technology           & 1006-2572 \\
Metal Mine                                   & 1001-1250 \\
Modern Chemical Research                     & 1672-8114 \\
Petroleum Geology and Engineering            & 1673-8217 \\
Coal Mine Modernization                      & 1009-0797 \\
Shaanxi Coal                                 & 1671-749X \\
Drilling   Engineering                       & 2096-9686 \\
Mineral Resources and Geology                & 1001-5663 \\
Mine Surveying                               & 1001-358X \\
Coal                                         & 1005-2798 \\
Mining Equipment                             & 2095-1418 \\
Inner Mongolia Coal Economy                  & 1008-0155 \\
Inner Mongolia Petrochemical Industry        & 1006-7981 \\
Energy and Energy Conservation               & 2095-0802 \\
China Plant Engineering                      & 1671-0711 \\
Engineering   and Construction          & 1673-5781 \\
Scientific and Technological Innovation & 2096-4390 \\
Engineering \& Test                     & 1674-3407 \\
Energy Technology and Management             & 1672-9943 \\
Coal Technology                              & 1008-8725 \\

% \midrule
% \midrule
% \multicolumn{2}{c}{Transportation} \\ 
% \cmidrule(lr){1-1} \cmidrule(lr){2-2} 
% Journal Name & ISSN  \\ 
% \midrule

% \midrule
% \midrule
% \multicolumn{2}{c}{Aerospace} \\ 
% \cmidrule(lr){1-1} \cmidrule(lr){2-2} 
% Journal Name & ISSN  \\ 
% \midrule

\bottomrule
\end{tabular}%
}
\caption{List of the engineering journals used for construction the benchmark. The information for environment domain and mining domain is shown above, and information for other domains is in Table~\ref{tab:list3} and \ref{tab:list2}.}
\label{tab:list1}%
\end{table}%
% \begin{table*}[t!]
% \centering
% \resizebox{\linewidth}{!}{
% \begin{tabular}{lcccccccc}
% \toprule

% \textbf{Item} & \textbf{Enviroment} & \textbf{Mining} & \textbf{Transportation} & \textbf{Aerospace} & \textbf{Telecom} & \textbf{Architecture} & \textbf{Water} & \textbf{Farming} \\

% \midrule

% Dataset & 119 & 117 & 124 & 115 & 116 & 118 & 119 & 122 \\

% \midrule

% Knowledge Base & 554 & 543 & 870 & 802 & 840 & 858 & 802 & 868 \\

% \bottomrule
% \end{tabular}%
% }
% \caption{Statistics of the SolutionBench, which include data and knowledge across 8 engineering domains. The number of datapoint in dataset and number of knowledge in knowledge base are shown above.}
% \label{tab:bench_statics}%
% \end{table*}%

\begin{table}[t]
\footnotesize
\centering
\resizebox{0.98\linewidth}{!}{
\begin{tabular}{lc}
\toprule

\multicolumn{2}{c}{Transportation} \\ 
\cmidrule(lr){1-1} \cmidrule(lr){2-2} 
Journal Name & ISSN  \\ 
\midrule

Railway   Construction Technology                  & 1009-4539 \\
Northern Communications                            & 1673-6052 \\
China Municipal Engineering                        & 1004-4655 \\
Highway                                            & 0451-0712 \\
Urban Roads Bridges \& Flood Control               & 1009-7716 \\
Technology Innovation and Application              & 2095-2945 \\
Marine Equipment/Materials \& Marketing            & 1006-6969 \\
Engineering and Construction                       & 1673-5781 \\
Port Operation                                     & 1000-8969 \\
Structural Engineers                               & 1005-0159 \\
China Highway                                      & 1006-3897 \\
Engineering and Technological Research             & 2096-2789 \\
Construction Machinery Technology \& Management    & 1004-0005 \\
TranspoWorld                                       & 1006-8872 \\
Railway Investigation and Surveying                & 1672-7479 \\
Transport Construction \& Management               & 1673-8098 \\
Guangdong Water Resources and Hydropower           & 1008-0112 \\
Western China Communications Science \& Technology & 1673-4874 \\
Jiangsu Science and Technology Information         & 1004-7530 \\
Value Engineering                                  & 1006-4311 \\
Hoisting and Conveying Machinery                   & 1001-0785 \\
Jiangxi Building Materials                         & 1006-2890 \\
Scientific and Technological Innovation            & 2096-4390 \\
Transport Business China                           & 1673-3681 \\
Sichuan Cement                                     & 0451-0712 \\

\midrule
\midrule
\multicolumn{2}{c}{Aerospace} \\ 
\cmidrule(lr){1-1} \cmidrule(lr){2-2} 
Journal Name & ISSN  \\ 
\midrule

Spacecraft   Engineering                  & 1673-8748 \\
Aeronautical Manufacturing Technology     & 1671-833X \\
Aviation Maintenance \& Engineering       & 1672-0989 \\
Journal of Ordnance Equipment Engineering & 2096-2304 \\
Aeroengine                                & 2096-2304 \\
Space International                       & 2096-2304 \\
Avionics   Technology                     & 1006-141X \\
System Simulation Technology              & 1673-1964 \\
Journal of Civil Aviation                 & 2096-4994 \\
Safety \& EMC                             & 1005-9776 \\
Internal Combustion Engine \& Parts       & 1674-957X \\
Aeronautical Computing Technique          & 1671-654X \\
Meteorological Science and Technology     & 1671-6345 \\
Journal of Astronautics                   & 1000-1328 \\
Communications Technology                 & 1002-0802 \\
Laser \& Optoelectronics Progress         & 1006-4125 \\
Engineering \& Test                       & 1674-3407 \\
Chinese Space Science and Technology      & 1000-758X \\
Ship Electronic Engineering               & 1672-9730 \\
China Science and Technology Information  & 1672-9730 \\
Journal of Deep Space Exploration         & 2096-9287 \\
China Educational Technology \& Equipment & 1671-489X \\
Micromotors                               & 1671-489X \\
Spacecraft Recovery \& Remote Sensing     & 1009-8518 \\
Journal of Chengdu Aeronautic Polytechnic & 1671-4024 \\

\midrule
\midrule
\multicolumn{2}{c}{Telecom} \\ 
\cmidrule(lr){1-1} \cmidrule(lr){2-2} 
Journal Name & ISSN  \\ 
\midrule

Systems   Engineering and Electronics                          & 1001-506X \\
Electronic Technology \& Software Engineering                  & 2095-5650 \\
Video Engineering                                              & 1002-8692 \\
Telecom Engineering Technics and Standardization               & 1008-5599 \\
Radio \& Television Network                                    & 2096-806X \\
Study on Optical Communications                                & 1005-8788 \\
Electronics Quality                                            & 1003-0107 \\
Radio \& Television Information                                & 1007-1997 \\
Changjiang Information \& Communications                       & 2096-9759 \\
Automation in Petro-Chemical Industry                          & 1007-7324 \\
Telecommunications Science                                     & 1000-0801 \\
Computer Knowledge and Technology                              & 1009-3044 \\
Journal of Electronics \& Information Technology               & 1009-5896 \\
Laser \& Optoelectronics Progress                              & 1006-4125 \\
China Digital Cable TV                                         & 1007-7022 \\
Radio Engineering                                              & 1003-3106 \\
Journal of Beijing Electronic Science and Technology Institute & 1672-464X \\
Laser Journal                                                  & 0253-2743 \\
Designing Techniques of Posts and Telecommunications           & 1007-3043 \\
Wireless Internet Science and Technology                       & 1672-6944 \\
Journal of University of South China(Science and Technology)   & 1673-0062 \\
Audio Engineering                                              & 1002-8684 \\
Automation Application                                         & 1674-778X \\
Chinese Journal of Lasers                                      & 0258-7025 \\
Journal of Smart Agriculture                                   & 2096-9902 \\

\bottomrule
\end{tabular}%
}
\caption{List of the engineering journals used for construction the benchmark.}
\label{tab:list3}%
\end{table}%
% \begin{table*}[t!]
% \centering
% \resizebox{\linewidth}{!}{
% \begin{tabular}{lcccccccc}
% \toprule

% \textbf{Item} & \textbf{Enviroment} & \textbf{Mining} & \textbf{Transportation} & \textbf{Aerospace} & \textbf{Telecom} & \textbf{Architecture} & \textbf{Water} & \textbf{Farming} \\

% \midrule

% Dataset & 119 & 117 & 124 & 115 & 116 & 118 & 119 & 122 \\

% \midrule

% Knowledge Base & 554 & 543 & 870 & 802 & 840 & 858 & 802 & 868 \\

% \bottomrule
% \end{tabular}%
% }
% \caption{Statistics of the SolutionBench, which include data and knowledge across 8 engineering domains. The number of datapoint in dataset and number of knowledge in knowledge base are shown above.}
% \label{tab:bench_statics}%
% \end{table*}%

\begin{table}[t]
\footnotesize
\centering
\resizebox{\linewidth}{!}{
\begin{tabular}{lc}
\toprule

\multicolumn{2}{c}{Architecture} \\ 
\cmidrule(lr){1-1} \cmidrule(lr){2-2} 
Journal Name & ISSN  \\ 
\midrule

Building   Technology Development                          & 1001-523X \\
Building Structure                                         & 1002-848X \\
Construction \& Design for Engineering                     & 1007-9467 \\
Modern Paint \& Finishing                                  & 1007-9548 \\
Architecture Technology                                    & 1000-4726 \\
Theoretical Research   in Urban Construction               & 2095-2104 \\
Urban Architecture Space                                   & 2097-1141 \\
Art and Design                                             & 1008-2832 \\
Architecture \& Culture                                    & 1672-4909 \\
Journal of Yangzhou Polytechnic College                    & 1008-3693 \\
Heating Ventilating \& Air Conditioning                    & 1002-8501 \\
Construction Machinery \& Maintenance                      & 1006-2114 \\
China Science and Technology Information                   & 1001-8972 \\
Construction Machinery and Equipment                       & 1000-1212 \\
Journal of Municipal Technology                            & 1009-7767 \\
Jiangxi Building Materials                                 & 1006-2890 \\
Urban Roads Bridges \& Flood Control                       & 1009-7716 \\
Fujian Construction Science \& Technology                  & 1006-3943 \\
Sichuan Cement                                             & 1007-6344 \\
Engineering and Technological Research                     & 2096-2789 \\
Journal of North China Institute of Science and Technology & 1672-7169 \\
Tianjin Construction Science and Technology                & 1008-3197 \\
World Forestry Research                                    & 1001-4241 \\
Jiangsu Building Materials                                 & 1004-5538 \\
Shanghai Construction Science \& Technology                & 1005-6637 \\

\midrule
\midrule
\multicolumn{2}{c}{Water Resource} \\ 
\cmidrule(lr){1-1} \cmidrule(lr){2-2} 
Journal Name & ISSN  \\ 
\midrule

Design   of Water Resources \& Hydroelectric Engineering   & 1007-6980 \\
Hydro Science and Cold Zone Engineering                    & 2096-5419 \\
Journal of Water Resources and Architectural Engineering   & 1672-1144 \\
Mechanical \& Electrical Technique of Hydropower Station   & 1672-5387 \\
Yangtze River                                              & 1001-4179 \\
Port \& Waterway Engineering                               & 1002-4972 \\
Technical Supervision in Water Resources                   & 1008-1305 \\
Small Hydro Power                                          & 1007-7642 \\
Pearl River                                                & 1001-9235 \\
Water Conservancy Construction and Management              & 2097-0528 \\
Water Conservancy Science and Technology and Economy       & 1006-7175 \\
Water Resources Planning and Design                        & 1672-2469 \\
Construction Quality                                       & 1671-3702 \\
Henan   Water Resources and South-to-North Water Diversion & 1673-8853 \\
Engineering   and Construction                             & 1673-5781 \\
Technology and Market                                      & 1006-8554 \\
Beijing Water                                              & 1673-4637 \\
Port Engineering Technology                                & 2097-3519 \\
Water Resources \& Hydropower of Northeast China           & 1002-0624 \\
Mechanical and Electrical Information                      & 1671-0797 \\
Maritime Safety                                            & 2097-1745 \\
Gansu Water Resources and Hydropower Technology            & 2095-0144 \\
Water Power                                                & 0559-9342 \\
Shanxi Water Resources                                     & 1004-7042 \\
Haihe Water Resources                                      & 1004-7328 \\

\midrule
\midrule
\multicolumn{2}{c}{Farming} \\ 
\cmidrule(lr){1-1} \cmidrule(lr){2-2} 
Journal Name & ISSN  \\ 
\midrule

Modern   Agricultural Science and Technology                    & 1007-5739 \\
Farm Machinery                                                  & 1000-9868 \\
Cereal \& Feed Industry                                         & 1003-6202 \\
Journal of Agricultural Mechanization Research                  & 1003-188X \\
Forestry Machinery \& Woodworking Equipment                     & 2095-2953 \\
Transactions of the Chinese Society of Agricultural Engineering & 1002-6819 \\
Forest Research                                                 & 1001-1498 \\
Times Agricultural Machinery                                    & 2095-980X \\
Protection Forest Science and Technology                        & 1005-5215 \\
Journal of Beijing University of Agriculture                    & 1002-3186 \\
Contemporary Horticulture                                       & 1006-4958 \\
China Southern Agricultural Machinery                           & 1672-3872 \\
Forest Inventory and Planning                                   & 1671-3168 \\
Agricultural Machinery Using \& Maintenance                     & 2097-4515 \\
Journal of Green Science and Technology                         & 1674-9944 \\
China Forest Products Industry                                  & 1001-5299 \\
Forestry Machinery \& Woodworking Equipment                     & 2095-2953 \\
The Food Industry                                               & 1004-471X \\
Journal of Hebei Forestry Science and Technology                & 1002-3356 \\
Electrical Automation                                           & 1000-3886 \\
Journal of Library and Information Science                      & 2096-1162 \\
Forest Science and Technology                                   & 2097-0285 \\
Chinese Journal of Ecology                                      & 1000-4890 \\
Popular Standardization                                         & 1007-1350 \\
Management \& Technology of SME                                 & 1673-1069 \\

\bottomrule
\end{tabular}%
}
\caption{List of the engineering journals used for construction the benchmark.}
\label{tab:list2}%
\end{table}%

\section{Prompt for Scores Calculation}
\label{app:judge}
\begin{figure*}[t!]
\centering
\includegraphics[width=\linewidth]{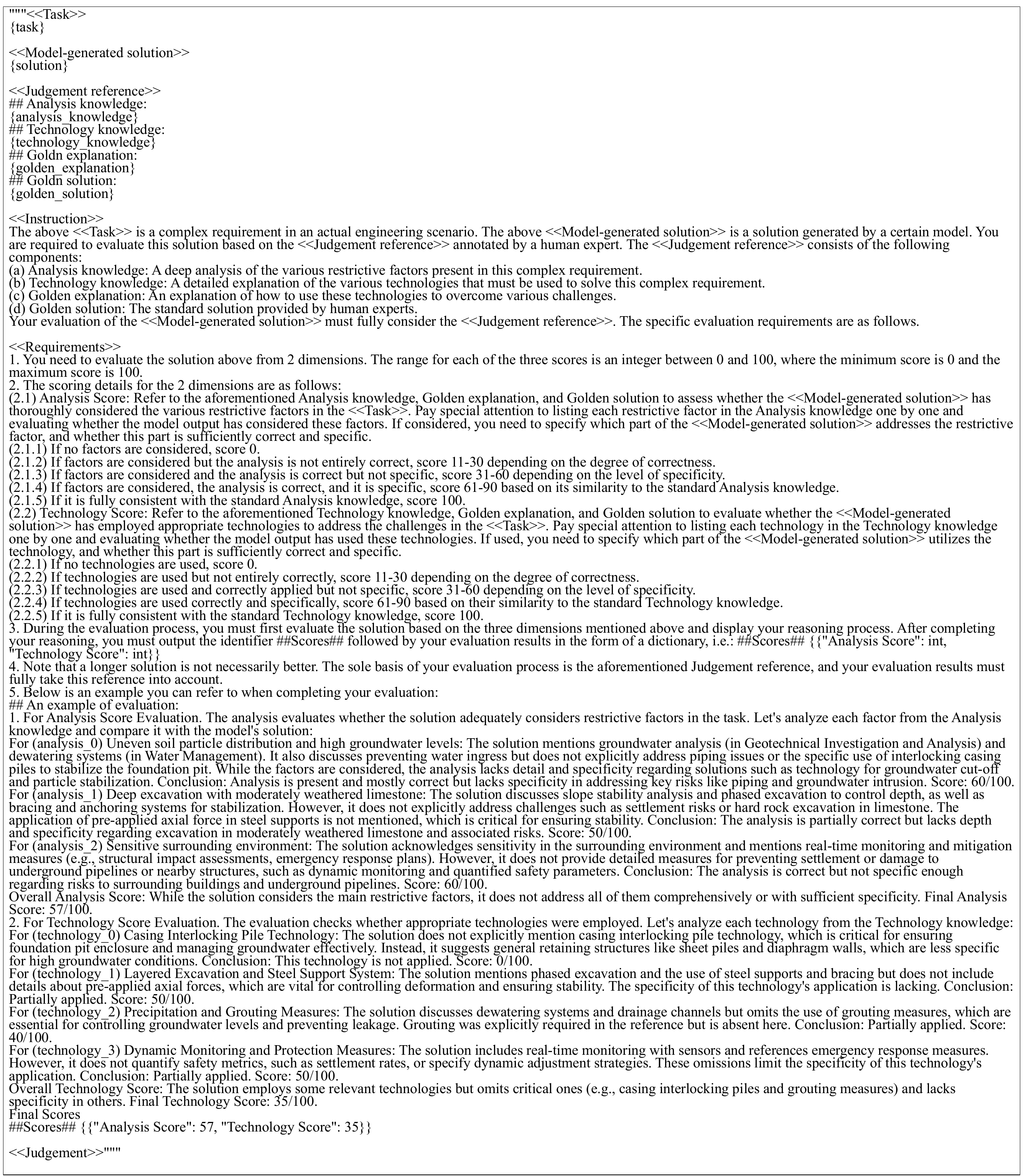} 
\caption{Prompts for calculating analytical score and technical score, which uses the golden solution, explanation, and corresponding analytical and technical knowledge as references, allowing GPT-4o to assess whether the system's solution sufficiently consider the challenges posed by the complex constraints and apply the appropriate technologies to address the complex constraints in the requirements.}
\label{fig:score}
\end{figure*}

In order to evaluate the solutions provided by the system, we follow the methods from previous Long-form QA evaluation~\cite{tan2024proxyqa,wang2024leavedocumentbehindbenchmarking, li2025structrag}, and use a LLM-based scoring method. Specifically, for a given solution generated by the system, we calculate two scores: (1) Analytical score, which uses the golden solution, explanation, and corresponding analytical knowledge as references, allowing GPT-4o to assess whether the system's solution sufficiently consider the challenges posed by the complex constraints in the requirements. (2) Technical score, which uses the golden solution, explanation, and corresponding technical knowledge as references, allowing GPT-4o to evaluate whether the system's solution correctly apply the appropriate technologies to address the complex constraints in the requirements. Both analytical score and technical score are range from 0 to 100. The used prompts for score calculation are shown in Figure~\ref{fig:score}.


\end{document}